\documentclass[10pt,twocolumn,letterpaper]{article}
\usepackage{cvpr}              
\usepackage{amsmath}
\usepackage{amssymb}
\usepackage{bbm}
\usepackage{booktabs}
\usepackage{color}
\usepackage{dsfont}
\usepackage{graphicx}
\usepackage{multirow}
\usepackage{pifont}
\usepackage[pagebackref,breaklinks,colorlinks]{hyperref}
\usepackage[capitalize]{cleveref}

\definecolor{myGreen}{rgb}{0, .6, .0}
\hypersetup{colorlinks,breaklinks,anchorcolor=darkblue,citecolor=myGreen}

\crefname{section}{Sec.}{Secs.}
\Crefname{section}{Section}{Sections}
\Crefname{table}{Table}{Tables}
\crefname{table}{Tab.}{Tabs.}

\newcommand{\cmark}{\ding{51}}%
\newcommand{\xmark}{\ding{55}}%

\def\cvprPaperID{9552}
\def\confName{CVPR}
\def\confYear{2022}

\usepackage{overpic}
\usepackage{enumitem} 
\usepackage{overpic} 
\usepackage{color}

\definecolor{turquoise}{cmyk}{0.65,0,0.1,0.3}
\definecolor{purple}{rgb}{0.65,0,0.65}
\definecolor{dark_green}{rgb}{0, 0.5, 0}
\definecolor{orange}{rgb}{0.8, 0.6, 0.2}
\definecolor{red}{rgb}{1.0, 0.0, 0.0}
\definecolor{darkred}{rgb}{0.6, 0.1, 0.05}
\definecolor{blueish}{rgb}{0.0, 0.3, .6}
\definecolor{light_gray}{rgb}{0.7, 0.7, .7}
\definecolor{pink}{rgb}{1, 0, 1}
\definecolor{greyblue}{rgb}{0.25, 0.25, 1}

\newcommand{\todo}[1]{{\color{black}#1}}

\usepackage{blindtext}

\makeatletter
\renewcommand{\paragraph}{%
  \@startsection{paragraph}{4}%
  {\z@}{0.215em}{-1em}%
  {\normalfont\normalsize\bfseries}%
}
\makeatother

\begin{document}
\title{Generalized Category Discovery}

\author{Sagar Vaze$^\star$ \qquad Kai Han$^\dagger$ \qquad Andrea Vedaldi$^\star$ \qquad Andrew Zisserman$^\star$\\
\small 
$^\star$Visual Geometry Group, Department of Engineering Science, University of Oxford \\
\small
$^\dagger$The University of Hong Kong\\
{\tt\small \{sagar,vedaldi,az\}@robots.ox.ac.uk \qquad kaihanx@hku.hk}
}
\twocolumn[{%
\renewcommand\twocolumn[1][]{#1}%
\maketitle
\begin{center}
\centering
\captionsetup{type=figure}
\includegraphics[width=\textwidth]{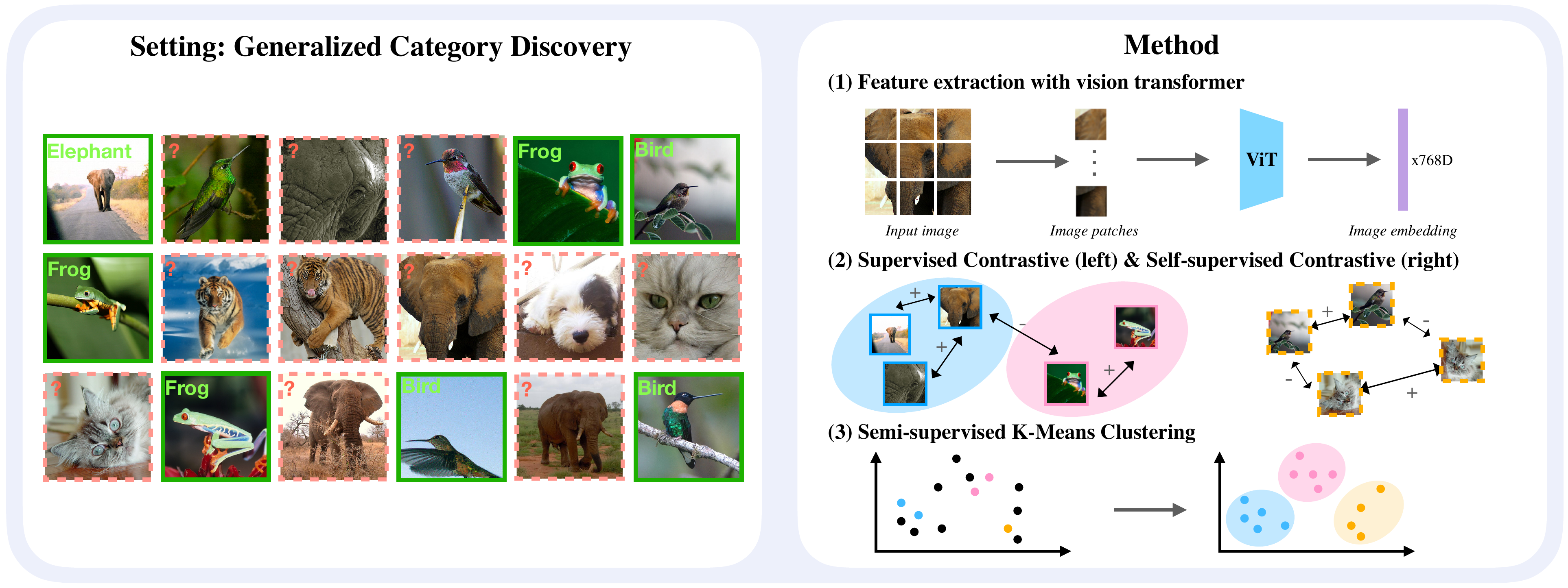}
\captionof{figure}{We present a new setting: `Generalized Category Discovery' and a method to tackle it. Our setting can be succinctly described as: given a dataset, a subset of which has class labels, categorize all unlabelled images in the dataset. The unlabelled images may come from labelled or novel classes. Our method leverages contrastively trained vision transformers to assign labels directly through clustering.}\label{fig:overview}
\end{center}
}]
\begin{abstract}
In this paper, we consider a highly general image recognition setting wherein, given a labelled and unlabelled set of images, the task is to categorize all images in the unlabelled set.
Here, the unlabelled images may come from labelled classes or from novel ones.
Existing recognition methods are not able to deal with this setting, because they make several restrictive assumptions, such as the unlabelled instances only coming from known -- or unknown -- classes, and the number of unknown classes being known a-priori. 
We address the more unconstrained setting, naming it `Generalized Category Discovery', and challenge all these assumptions. 
We first establish strong baselines by taking state-of-the-art algorithms from novel category discovery and adapting them for this task.
Next, we propose the use of vision transformers with contrastive representation learning for this open-world setting. 
We then introduce a simple yet effective semi-supervised $k$-means method to cluster the unlabelled data into seen and unseen classes automatically, substantially outperforming the baselines. 
Finally, we also propose a new approach to estimate the number of classes in the unlabelled data. 
We thoroughly evaluate our approach on public datasets for generic object classification and on fine-grained datasets, leveraging the recent Semantic Shift Benchmark suite. Code: \href{https://www.robots.ox.ac.uk/~vgg/research/gcd/}{https://www.robots.ox.ac.uk/$\sim$vgg/research/gcd}
\end{abstract}
\section{Introduction}\label{sec:intro}

Consider an infant sitting in a car and observing the world.
Object instances will pass the car and, for some of these, the infant may have been told their category (`that is a dog', `that is a car') and be able to recognize them.
There will also be instances that the infant has not seen before (cats and bicycles) and, having seen a number of these instances, we might expect the infant's visual recognition system to cluster these into new categories.

This is the problem that we consider in this work:
given an image dataset where only some images are labelled with their categories, assign a category label to each of the rest images, possibly using new categories not observed in the labelled set.
We term this problem \textit{Generalized Category Discovery} (GCD), and suggest that this is a realistic use case for many machine vision applications: whether that is recognizing products in a supermarket; pathologies in medical images; or vehicles in autonomous driving. 
\todo{In these and other realistic vision settings, it is often impossible to know if new images come from labelled or novel categories.}

In contrast, consider the limitations of existing image recognition settings.
In image classification, one of the most widely studied problems, all of the training images come with class labels.
Furthermore, all images at test time come from the same classes as the training set.
Semi-supervised learning (SSL)~\cite{chapelle2006semi} introduces the problem of learning from unlabelled data, but still assumes that all unlabelled images come from the same set of classes as the labelled ones.
More recently, the tasks of open-set recognition (OSR)~\cite{Scheirer_2013_TPAMI} and novel-category discovery (NCD)~\cite{han2019learning} have tackled open-world settings in which the images at test time may belong to new classes.
However, OSR aims only to \textit{detect} test-time images which do not belong to one of the classes in the labelled set, but does not require any further classification amongst these detected images.
Meanwhile, in NCD, the closest setting to the one tackled in this work, methods learn from labelled and unlabelled images, and aim to discover new classes in the unlabelled set.
However, NCD still makes the limiting assumption that \textit{all} of the unlabelled images come from new categories,  which is usually unrealistic.

In this paper, we tackle Generalized Category Discovery in a number of ways.
Firstly, we establish strong baselines by taking representative methods from NCD and applying them to this task. 
To do this, we adapt their training and inference mechanisms to account for our more general setting, as well as retrain them with a more robust backbone architecture.
We show that existing NCD methods are prone to overfit the labelled classes in this generalized setting. 

Next, observing the potential for NCD methods to overfit their classification heads to the labelled classes, we propose a simple but effective method for recognition by clustering.
Our key insight is to leverage the strong `nearest neighbour' classification property of vision transformers along with contrastive learning.
We propose the use of contrastive training and a semi-supervised $k$-means clustering algorithm to recognize images without a parametric classifier.
We show that these proposed methods substantially outperform the established baselines, both on generic object recognition datasets and, particularly, on more challenging fine-grained benchmarks.
For the latter evaluations, we leverage the recently proposed Semantic Shift Benchmark suite \cite{vaze2022openset}, which was designed for the task of identifying semantic novelty.

Finally, we propose a solution to a challenging and under-investigated problem in image recognition: estimating the number of categories in unlabelled data.
Almost all methods, including purely unsupervised ones, assume the knowledge of the number of categories, a highly unrealistic assumption in the real world.
We propose an algorithm which leverages the labelled set to tackle this problem.

\todo{Our contributions can be summarized as follows:
(i) the formalization of Generalized Category Discovery (GCD), a new and realistic setting for image recognition;
(ii) the establishment of strong baselines by adapting state-of-the-art techniques from standard novel category discovery to this task;
(iii) a simple but effective method for GCD, which uses contrastive representation learning and clustering to directly provide class labels, and outperforms the baselines substantially;
(iv) a novel method for estimating the number of categories in unlabelled data, a largely understudied problem;
and (v) rigorous evaluation on standard image recognition datasets as well as the recent Semantic Shift Benchmark suite \cite{vaze2022openset}. 
}



\section{Related work}
\label{sec:related}

Our work relates to prior work on \textit{semi-supervised learning}, \textit{open-set recognition}, and \textit{novel category discovery}, which we briefly review next.

\paragraph{Semi-supervised learning}
A number of methods~\cite{chapelle2006semi,oliver2018realistic,rebuffi2019semi,sohn2020fixmatch,48416} have been proposed to tackle the problem of semi-supervised learning (SSL). 
SSL assumes that the labelled and unlabelled instances come from the same set of classes. 
The objective is to learn a robust classification model leveraging both the labelled and unlabelled data during training. 
Amongst existing methods, consistency based approaches appear to be popular and effective, such as LadderNet~\cite{rasmus2015semi}, PI model~\cite{laine2016temporal}, Mean-teacher~\cite{tarvainen2017mean}. Recently, with the success of self-supervised learning, methods have also been proposed to improve SSL by augmenting the methods with self-supervised objectives~\cite{rebuffi2019semi,48416}. 

\paragraph{Open-set recognition}
The problem of open-set recognition (OSR) is formalized in~\cite{Scheirer_2013_TPAMI}, with the objective being to classify unlabelled instances from the same semantic classes as the labelled data, while detecting test instances from unseen classes. 
OpenMax~\cite{BendaleB15} is the first deep learning method to approach this problem with Extreme Value Theory. GANs are often employed to generate adversarial samples to train an open-set classifier, e.g, \cite{Neal_2018_ECCV,Ge2017Generative,kong2021}. Several methods have been proposed to train models such that images with large reconstruction error are regarded as open-set samples~\cite{Yoshihashi_2019_CVPR,Oza2019C2AE,Sun_2020_CVPR}. There are also methods that learns prototypes for the labelled classes, and identify the images from unknown classes by the distances to the prototypes~\cite{Shu2020podn,Chen_2020_ECCV,chen2021adversarial}.  
More recently, \cite{Chen_2020_ECCV,chen2021adversarial} proposed to learn reciprocal points which describe `otherness' with respect to the labelled classes. 
\cite{openhybrid20} jointly trains a flow-based density estimator and a classification based encoder for OSR. Finally, Vaze et al. \cite{vaze2022openset} study the correlation between the closed-set and open-set performance, showing that state-of-the-art OSR results can be obtained by boosting the closed-set accuracy of the standard cross-entropy baseline.

\paragraph{Novel category discovery}
The problem of novel category discovery (NCD) is formalized in DTC~\cite{han2019learning}. Earlier methods that could be applied to this problem include KCL~\cite{Hsu18_L2C} and MCL~\cite{Hsu19_MCL}, both of which maintain two models trained with labelled data and unlabelled data respectively, for general task transfer learning. 
AutoNovel (aka Rankstats)~\cite{han21autonovel,han20automatically} tackles the NCD problem with a three stage method. 
The model is first trained with self-supervision on all data for low-level representation learning. 
Then, it is further trained with full supervision on labelled data to capture higher level semantic information. Finally, a joint learning stage is carried out to transfer knowledge from the labelled to unlabelled data with ranking statistics.
Zhao and Han~\cite{zhao21novel} propose a model with two branches, one for global feature learning and the other for local feature learning, such that dual ranking statistics and mutual learning are conducted with these two branches for better representation learning and new class discovery.
OpenMix~\cite{zhong2021openmix} mixes the labelled and unlabelled data to avoid the model from over-fitting for NCD. NCL~\cite{Zhong_2021_CVPR} extracts and aggregates the pairwise pseudo-labels for the unlabelled data with contrastive learning and generates hard negatives by mixing the labelled and unlabelled data in the feature space for NCD.
Jia~et al.~\cite{jia21joint} propose an end-to-end NCD method for single- and multi-modal data with contrastive learning and winner-takes-all hashing.
A unified cross-entropy loss is introduced in UNO~\cite{Fini_2021_ICCV} to allow the model to be trained on labelled and unlabelled data jointly, by swapping the pseudo-labels from labelled and unlabelled classification heads.

\todo{Finally, we highlight the work by Girish et al. \cite{girish2020attribution} that tackles a similar setting to GCD but for the task of \textit{GAN attribution} instead of image recognition, as well as the concurrent work by Cao et al. \cite{cao2022openworld} that tackles a similar setting for image recognition under the name \textit{Open World Semi-Supervised Learning}. 
Different to our setting, they do not leverage large-scale pretraining or demonstrate performance on the Semantic Shift Benchmark, which better isolates the problem of detecting semantic novelty.} 
\section{Generalized category discovery}

We first formalize the task of \emph{Generalized Category Discovery} (GCD).
In short, we consider the problem of classifying images in a dataset, a subset of which has known class labels.
The task is to assign class labels to all remaining images, using classes that may or may not be observed in the labelled images (see \cref{fig:overview}, left).

Formally, we define GCD as follows.
We consider a dataset $\mathcal{D}$ comprising two parts
$\mathcal{D_L} = \{(\mathbf{x}_i, y_i)\}_{i=1}^{N} \in \mathcal{X} \times \mathcal{Y_L}$
and
$\mathcal{D_U} = \{(\mathbf{x}_i, y_i)\}_{i=1}^{M} \in \mathcal{X} \times \mathcal{Y_U}$,
where $\mathcal{Y_L} \subset \mathcal{Y_U}$.
During training, the model does not have access to the labels in $\mathcal{D_U}$, and is tasked with predicting them at test time.
Furthermore, we assume access to a validation set,
$\mathcal{D_V} = \{(\mathbf{x}_i, y_i)\}_{i=1}^{N'} \in \mathcal{X} \times \mathcal{Y_L}$,
which is disjoint from the training set and contains images from the same classes as the labelled set.
This formalization allows us to clearly see the distinction with the novel category discovery setting.
NCD assumes $\mathcal{Y_L} \cap \mathcal{Y_U} = \emptyset$ and existing methods rely on this prior knowledge during training.

In this section, we describe the methods we propose to tackle GCD\@.
First, we describe our approach to the problem.
Leveraging recent progress in self-supervised representation learning, we propose a simple but effective approach based on contrastive learning, with classification performed by a semi-supervised $k$-means algorithm.
Next, we develop a method to estimate the number of categories in the unlabelled data -- a challenging task that is understudied in the literature.
Finally, we build two strong baselines for GCD by modifying state-of-the-art NCD methods, RankStats~\cite{han21autonovel} and UNO~\cite{Fini_2021_ICCV}, to fit with our setting.

\subsection{Our approach}

The key insight of our approach for image recognition in an open-world setting is to remove the need for parametric classification heads.
Instead, we perform clustering directly in the feature space of a deep network (see \cref{fig:overview}, right).
Classification heads (typically, linear classifiers on top of a learned embedding) are best trained with the cross-entropy loss, which has been shown to be susceptible to noisy labels~\cite{Lei20can}.
Furthermore, when training a linear classifier for unlabelled classes, a typical method is to generate (noisy) pseudo-labels for the unlabelled instances.
This would suggest that parametric heads are susceptible to performance deterioration on the unlabelled classes.
Finally, we note that, by necessity, classification heads must be trained from scratch, which further makes them vulnerable to overfitting on the labelled classes.

Meanwhile, self-supervised contrastive learning has been widely used as pre-training to achieve robust representations in NCD~\cite{jia21joint,Zhong_2021_CVPR}.
Furthermore, when combined with vision transformers, it generates models which are good nearest neighbour classifiers~\cite{caron2021emerging}.
Inspired by this, we find that constrastively training a ViT model allows us to directly cluster in the model's feature space, thereby removing the need for a linear head which could lead to overfitting.
Specifically, we train the representation with a noise contrastive loss~\cite{gutmann2010noise} on all images \emph{without using any labels}.
This is important because it avoids overfitting the features to the subset of classes that are (partially) labelled.
We add a further supervised contrastive component~\cite{khosla2020supervised} for the labelled instances to make use of the labelled data (see \cref{fig:overview}, middle row on the right).

\subsubsection{Representation learning}

We use, for all methods, a vision transformer (ViT-B-16)~\cite
{dosovitskiy2021an} pretrained with DINO~\cite{caron2021emerging} self-supervision on (unlabelled) ImageNet~\cite{deng09imagnet} as our backbone.
This is motivated firstly because the DINO model is a strong nearest neighbour classifier, which suggests non-parametric clustering in its feature space would work well.
Secondly, self-supervised vision transformers have demonstrated the attractive quality of learning to attend to salient parts of an object without human annotation.
We find this feature to be useful for this task, because \emph{which} object parts are important for classification is likely to transfer well from the labelled to the unlabelled categories (see \cref{sec:qualitative_results}).

Finally, we wish to reflect a realistic and practical setting.
In the NCD literature, it is standard to train a ResNet-18~\cite{he2016deep} backbone from scratch for the target task.
However, in a real-world setting, a model is often initialized with large-scale pretrained weights to optimize performance (often ImageNet supervised pretraining).
In order to avoid conflicts with our experimental setting (which assumes a finite labelled set), we use self-supervised ImageNet weights.
To enhance the representation such that it is more tailored for the labelled and unlabelled data we have, we further fine-tune the representation on our target data jointly with supervised contrastive learning on the labelled data, and unsupervised contrastive learning on \textit{all} the data.

Formally, let $\mathbf{x}_i$ and $\mathbf{x}_{i}'$ be two views (random augmentations) of the same image in a mini-batch $B$.
The unsupervised contrastive loss is written as:
\begin{equation}
\label{eq:nce_unsup}
\mathcal{L}^{u}_{i} =
- \log \frac{\exp \left(\mathbf{z}_i \cdot \mathbf{z}_{i}'/ \tau\right)}{\sum_{n}  \mathds{1}_{[n \ne i]} \exp \left(\mathbf{z}_i \cdot \mathbf{z}_n / \tau\right)},
\end{equation}
where $\mathbf{z}_i = \phi(f(\mathbf{x}_i))$ and $\mathds{1}_{[n \ne i]}$ is an indicator function evaluating to 1 \emph{iff}  $n \ne i$, and $\tau$ is a temperature value. $f$ is the feature backbone, and $\phi$ is a multi-layer perceptron (MLP) projection head.

The supervised contrastive loss is written as:
\begin{equation}
\label{eq:nce_sup}
\medmuskip=2mu
\thickmuskip=3mu
\renewcommand\arraystretch{1.5}
\mathcal{L}^{s}_{i} =
- \frac{1}{|\mathcal{N}(i)|}  \sum_{q \in \mathcal{N}(i)}\log \frac{\exp \left(\mathbf{z}_i \cdot \mathbf{z}_q / \tau\right)}{\sum_{n} \mathds{1}_{[n \ne i]} \exp \left(\mathbf{z}_i \cdot \mathbf{z}_n / \tau\right)},
\end{equation}
where $\mathcal{N}(i)$ denotes the indices of other images having the same label as $\mathbf{x}_i$ in the mini-batch $B$.
Finally, we construct the total loss over the batch as:
\begin{equation}
\label{eq:total_loss}
\mathcal{L}^{t} = (1 - \lambda) \sum_{i \in B} \mathcal{L}^{u}_{i} +
\lambda \sum_{i \in B_\mathcal{L}} \mathcal{L}^{s}_{i}
\end{equation}
where $B_\mathcal{L}$ corresponds to the labelled subset of $B$ and $\lambda$ is a weight coefficient.
Using the labels \textit{only} in a contrastive framework, rather than in a cross-entropy loss, means that unlabelled and labelled data are treated similarly.
The supervised contrastive component is only used to nudge the network towards a semantically meaningful representation, thereby minimizing overfitting on the labelled classes.

\subsubsection{Label assignment with semi-supervised $k$-means}
\label{sec:semi_sup_kmeans}

Given the learned representation for the data, we can now assign class or cluster labels for each unlabelled data point, either from the labelled classes or unseen new classes.
Instead of performing this parametrically as is common in NCD (and risk overfitting to the labelled data) we propose to use a non-parametric method.
Namely, we propose to modify the classic $k$-means into a constraint algorithm by forcing the assignment of the instances in $\mathcal{D_L}$ to the correct cluster based on their ground-truth labels. Note, here we assume knowledge of the number of clusters, $k$. We tackle the problem of estimating this parameter 
in \cref{sec:k_estimation}.
The initial $|\mathcal{Y_L}|$ centroids for $\mathcal{D_L}$ are obtained based on the ground-truth class labels, and an additional 
$|\mathcal{Y_U}\setminus \mathcal{Y_L}|$
(number of new classes) initial centroids are obtained from $\mathcal{D_U}$ with $k$-means++~\cite{Arthur2008kmeanspp}, constrained on the centroids of $\mathcal{D_L}$.
During each centroid update and cluster assignment cycle, instances from the same class in $\mathcal{D_L}$ are always forced to have the same cluster assignment, while each instance in $\mathcal{D_U}$ can be assigned to any cluster based on the distance to different centroids. After the semi-supervised $k$-means converges, each instance in $\mathcal{D_U}$ can be assigned a cluster label. We provide a clear diagram of this in \cref{sec:semi_sup_kmeans_sup}.


\subsection{Estimating the class number in unlabelled data}
\label{sec:k_estimation}

Here, we tackle the problem of finding the number of classes in the unlabelled data.
In the NCD and unsupervised clustering settings, prior knowledge of the number of categories in the dataset is often assumed, but this is unrealistic in the real world given that the labels themselves are unknown.
To estimate the number of categories in $\mathcal{D_U}$, we leverage the information available in $\mathcal{D_L}$.
Specifically, we perform $k$-means clustering on the entire dataset, $\mathcal{D}$, before evaluating clustering accuracy on \textit{only} the labelled subset (see \cref{evaluation_protocol} for the metric's definition).

Clustering accuracy is evaluated by running the Hungarian algorithm~\cite{kuhn1955hungarian} to find the optimal assignment between the set of cluster indices and ground truth labels.
If the number of clusters is higher than the total number of classes, the extra clusters are assigned to the null set, and all instances assigned to those clusters are said to have been predicted incorrectly.
Conversely, if the number of clusters is lower than the number of classes, extra classes are assigned to the null set, and all instances with those ground truth labels are said to have been to be predicted incorrectly.
Thus, we assume that, if the clustering (across $\mathcal{D}$) is performed with $k$ too high or too low, then this will be reflected in a sub-optimal clustering accuracy on $\mathcal{D_L}$.
In other words, we assume clustering accuracy on the labelled set will be maximized when $k = |\mathcal{Y_L} \cup \mathcal{Y_U}|$.
This intuition leads us to use the clustering accuracy as a `black box' scoring function, $ACC = f(k; \mathcal{D})$, which we optimize with Brent's algorithm to find the optimal $k$.
\todo{Different to the method in \cite{han21autonovel}, which exhaustively iterates through all possible values of $k$, we find that the black-box optimization allows our method to scale to datasets with many categories.

Finally, we highlight that labelled sets with different granularities would induce different estimates of the number of classes. 
However, we suggest that the labelled set defines the system of categorization --- that the granularity of a real-world dataset is not an intrinsic property of the images, but rather a framework imposed by the labels. For instance, in Stanford Cars, the dataset could equally be labelled at the ‘Manufacturer’, ‘Model’ or ‘Variant’ level, with the categorization system defined by the labels assigned.}

\subsection{Two strong baselines}

We adapt two methods from the nearest image recognition sub-field, novel category discovery (NCD), for our generalized category discovery (GCD) task.
\textit{RankStats}~\cite{han21autonovel} is widely used as a competitive baseline for novel category discovery, while \textit{UNO}~\cite{Fini_2021_ICCV} is to the best of our knowledge the state-of-the-art method for NCD\@.

\paragraph{Baseline: RankStats+}
RankStats trains two classifiers on top of a shared feature representation:
the first head is fed instances from the labelled set and is trained with the cross-entropy loss, while the second head sees only instances from unlabelled classes (again, in the NCD setting, the labelled and unlabelled classes are disjoint).
In order to adapt RankStats to GCD, we train it with a single classification head for the total number of classes in the dataset.
We then train the first $|\mathcal{Y_L}|$ elements of the head with the cross-entropy loss, and train the \textit{entire} head with the binary cross-entropy loss with pseudo-labels.

\paragraph{Baseline: UNO+}
Similarly to RankStats, UNO is trained with classification heads for labelled and unlabelled data.
The model is then trained in a SwAV-like manner~\cite{caron2020unsupervised}.
First, multiple views (random augmentations) of a batch are generated and fed to the same model.
For the labelled images in the batch, the labelled head is trained with the cross-entropy loss using the ground truth labels.
For the unlabelled images, predictions (logits from the unlabelled head) are gathered for a given view and used as pseudo-labels with which to optimize the loss from \textit{other} views.
To adapt this mechanism, we simply concatenate both the labelled and unlabelled heads, thus allowing generated pseudo-labels for the unlabelled samples to belong to any class in the dataset.
\section{Experiments}

\subsection{Experimental setup}

\paragraph{Data}
We demonstrate results on six datasets in our proposed setting.
For each dataset, we take the training set and sample a set of classes for which we have labels during training.
We further sub-sample 50\% of the images from these classes to constitute the labelled set $\mathcal{D_L}$.
The remaining instances from these classes, along with all instances from the other classes, constitute $\mathcal{D_U}$.
We further construct the validation set for the labelled classes from the test or validation split of each dataset.

We first demonstrate results on three generic object recognition datasets: CIFAR10~\cite{Krizhevsky09cifar}, CIFAR100~\cite{Krizhevsky09cifar} and ImageNet-100~\cite{deng09imagnet}.
ImageNet-100 refers to the ImageNet dataset with 100 classes randomly subsampled.
\todo{These datasets establish the methods' performance on well-known datasets in the standard image recognition literature.

We further evaluate on the recently proposed Semantic Shift Benchmark \cite{vaze2022openset} (SSB, including CUB~\cite{cub200} and Stanford Cars~\cite{Cars196}), as well as on Herbarium19~\cite{Chuan19herbarium}.
SSB provides fine-grained evaluation datasets with clear `axes of semantic variation' and further provides categories for $\mathcal{D_U}$ which are delineated from $\mathcal{D_L}$ in a semantically coherent fashion.
Thus, the user can be confident that the recognition system is identifying new classes based on a true semantic signal, rather than simply responding to low-level distributional shifts in the data, as may be the case for generic object recognition datasets.
The long-tailed nature of Herbarium19 adds an additional challenge to the evaluation.
}


The fine-grained datasets further reflect many real-world use cases for image recognition systems, which are deployed in constrained environments with many similar objects (\eg products in a supermarket, traffic monitoring, or animal tracking in the wild).
In fact, the Herbarium19 dataset itself represents a real-world use case for GCD: while we are aware of roughly 400k species of plants, and estimate that there are around 80k yet to be discovered, it currently takes roughly 35 years from plant collection to plant species description if performed manually~\cite{Chuan19herbarium}.
We summarize the dataset splits used in our evaluations in~\cref{tab:datasets}, and provide more details in \cref{sec:dataset_details_sup}.

\begin{table}
\centering
\caption{
%
Datasets used in our experiments. We show the number of classes in the labelled and unlabelled sets ($|\mathcal{Y_L}|, |\mathcal{Y_U}|$), as well as the number of images ($|\mathcal{D_L}|, |\mathcal{D_U}|$).
} 
\label{tab:datasets}
\resizebox{\linewidth}{!}{ 
\begin{tabular}{lcccccc}
\toprule
     & CIFAR10 & CIFAR100 & ImageNet-100 & CUB  & SCars & Herb19 \\ 
\midrule
$|\mathcal{Y_L}|$ & 5       & 80       & 50           & 100  & 98    & 341    \\ 
$|\mathcal{Y_U}|$ & 10      & 100      & 100          & 200  & 196   & 683    \\ 
\midrule
$|\mathcal{D_L}|$ & 12.5k   & 20k      & 31.9k        & 1.5k & 2.0k  & 8.9k   \\ 
$|\mathcal{D_U}|$ & 37.5k   & 30k      & 95.3k        & 4.5k & 6.1k  & 25.4k  \\ 
\bottomrule
\end{tabular}
}
\end{table}

\paragraph{Evaluation protocol}\label{evaluation_protocol}
For each dataset, we train the models on $\mathcal{D}$ (without access to the ground truth labels in $\mathcal{D_U}$).
At test-time, we measure the clustering accuracy between the ground truth labels $y_i$ and the model's predictions $\hat{y}_i$ as:
\begin{equation}
    ACC = \max_{p \in \mathcal{P(Y_U)}}
    \frac{1}{M}
    \sum_{i=1}^{M}
    \mathds{1}\{y_i = p(\hat{y}_i)\}
\end{equation}
Here, $M = |\mathcal{D_U}|$ and $\mathcal{P(Y_U)}$ is the set of all permutations of the class labels in the unlabelled set.
\todo{Our main metric is $ACC$ on `All' instances, indicating image recognition accuracy across the entire unlabelled set $\mathcal{D_U}$.
We further report values for both the `Old' classes subset (instances in $\mathcal{D_U}$ belonging to classes in $\mathcal{Y_L}$) and `New' classes subset (instances in $\mathcal{D_U}$ belonging to classes in $\mathcal{Y_U} \setminus \mathcal{Y_L}$).

The maximum over the set of permutations is computed via the Hungarian optimal assignment algorithm~\cite{kuhn1955hungarian}.
Importantly, we compute the Hungarian assignment only once, across all categories $\mathcal{Y_U}$, and measure classification accuracy on `Old' and `New' subsets only afterwards.
The interaction between when the Hungarian assignment is performed, and the resultant $ACC$ on the subsets, can be un-intuitive and is elaborated upon in \cref{sec:appendix_eval_metrics_details}.
}


\paragraph{Implementation details}
All methods are trained with a ViT-B-16 backbone with DINO pre-trained weights, and use the output \texttt{[CLS]} token as the feature representation.
All methods were trained for 200 epochs, with the best model selected using accuracy on the validation set.
We fine-tune the final transformer block for all methods.

For our method, we fine-tune the final block of the vision transformer with an initial learning rate of 0.1 which we decay with a cosine annealed schedule.
We use a batch size of 128 and $\lambda = 0.35$ in the loss (see \cref{eq:total_loss}).
Furthermore, following standard practise in self-supervised learning, we project the model's output through a non-linear projection head before applying the contrastive loss.
We use the same projection head as in \cite{caron2021emerging} and discard it at test-time.
For the baselines from NCD, we follow the original implementations and learning schedules as far as possible, referring to the original papers for details \cite{han21autonovel,Fini_2021_ICCV}.

%
Finally, in order to estimate $k$, we run our $k$-estimation method on DINO features extracted from each of the considered benchmarks.
We run Brent's algorithm on a constrained domain for $k$, with the minimum set at $|\mathcal{Y_L}|$ and the maximum set at 1000 classes for all datasets.

\subsection{Comparison with the baselines}

We report results for all compared methods in~\cref{tab:generic} and~\cref{tab:fine_grained}.
As an additional baseline, we also report results when running $k$-means directly on top of raw DINO features (reported as $k$-means).
\todo{~\cref{tab:generic} presents results on the generic object recognition datasets, while ~\cref{tab:fine_grained} shows results on SSB and Herbarium19.
We further show results on the FGVC-Aircraft \cite{maji13fine-grained} evaluation from SSB in \cref{sec:appendix_fgvc_aircraft}.

Overall (across `All' instances in $\mathcal{D_U}$), our method outperforms RankStats+ and UNO+ baselines on the standard image recognition datasets by 9.3\% in absolute terms, and 11.5\% in proportional terms.
Meanwhile, on the more challenging fine-grained evaluations, our method outperforms the baselines by 8.9\% in absolute terms and 27.0\% in proportional terms.

We find that on categories with labelled examples (`Old' classes), the baselines which use parametric classifiers can outperform our method, but this is at the expense of $ACC$ on the `New' categories.
We also found that, if the baselines were trained for longer, they would begin to sacrifice $ACC$ on `Old' categories for $ACC$ on `New' ones, but that best overall performance was achieved using early stopping by monitoring the performance on the validation set.
}

\begin{table}[htb]
\footnotesize
\centering
\caption{Results on generic image recognition datasets.}\label{tab:generic}
\resizebox{\linewidth}{!}{ 
\begin{tabular}{lccccccccc}
\toprule
& \multicolumn{3}{c}{CIFAR10} & \multicolumn{3}{c}{CIFAR100} & \multicolumn{3}{c}{ImageNet-100
}\\
\cmidrule(rl){2-4}
\cmidrule(rl){5-7}
\cmidrule(rl){8-10}
Classes       & All      & Old     & New        & All       & Old       & New       & All       & Old       & New \\\midrule
$k$-means~\cite{MackQueen67_Kmeans}
& 83.6
& 85.7
& 82.5
& 52.0
& 52.2
& 50.8
& 72.7
& 75.5
& \textbf{71.3}
\\
RankStats+
&  46.8
&  19.2
&  60.5
&  58.2
&  77.6
&  19.3
&  37.1
&  61.6
&  24.8
\\
UNO+ 
&  68.6
&  \textbf{98.3}
&  53.8
&  69.5
&  \textbf{80.6}
&  47.2
&  70.3
&  \textbf{95.0}
&  57.9
\\
\midrule
Ours
&  \textbf{91.5}
&  97.9
&  \textbf{88.2}
&  \textbf{73.0}
&  76.2
&  \textbf{66.5}
&  \textbf{74.1}
&  89.8
&  66.3
\\
\bottomrule
\end{tabular}
}
\end{table}
\begin{table}[htb]
\footnotesize
\centering
\caption{Results on SSB \cite{vaze2022openset} and Herbarium19\cite{Chuan19herbarium}. }\label{tab:fine_grained}
\resizebox{\linewidth}{!}{ 
\begin{tabular}{lccccccccc}
\toprule
& \multicolumn{3}{c}{CUB} & \multicolumn{3}{c}{Stanford Cars} & \multicolumn{3}{c}{Herbarium19}\\
\cmidrule(rl){2-4}
\cmidrule(rl){5-7}
\cmidrule(rl){8-10}
Classes       & All      & Old     & New        & All       & Old       & New       & All       & Old       & New \\\midrule
$k$-means~\cite{MackQueen67_Kmeans}
& 34.3
& 38.9
& 32.1
& 12.8
& 10.6
& 13.8
& 12.9
& 12.9
& 12.8
\\
RankStats+
& 33.3
& 51.6
& 24.2
& 28.3
& 61.8
& 12.1
& 27.9
& \textbf{55.8}
& 12.8
\\
UNO+ 
& 35.1
& 49.0
& 28.1
& 35.5
& \textbf{70.5}
& 18.6
& 28.3
& 53.7
& 14.7
\\
\midrule
Ours
& \textbf{51.3}
& \textbf{56.6}
& \textbf{48.7}
& \textbf{39.0}
& 57.6
& \textbf{29.9}
& \textbf{35.4}
& 51.0
& \textbf{27.0}
\\
\bottomrule
\end{tabular}
}
\end{table}
\begin{figure*}[htb]
\begin{center}
\includegraphics[width=1.\linewidth]{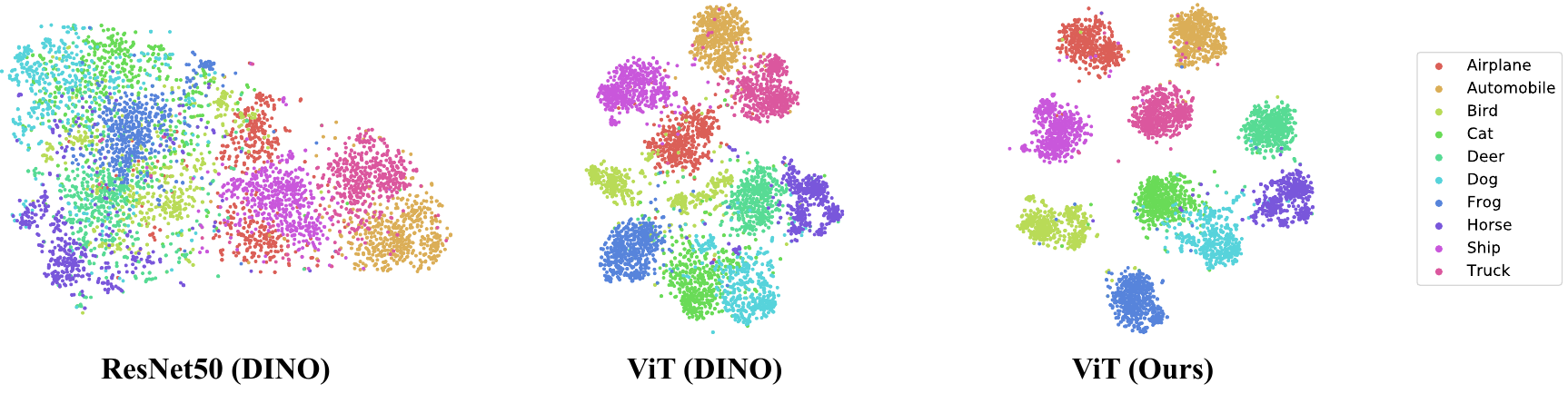}
\end{center}
\caption{TSNE visualization of instances in CIFAR10 for features generated by a ResNet-50 and ViT model trained with DINO self-supervision on ImageNet, and a ViT model after fine-tuning with our approach.}
\label{fig:tsne}
\end{figure*}
\begin{table}
\centering
\caption{
%
Estimation of the number of classes in unlabelled data.
} 
\label{tab:k_est}
\resizebox{\linewidth}{!}{ 
\begin{tabular}{@{}lcccccc@{}}
\toprule
   &CIFAR10 &CIFAR100 &ImageNet-100 & CUB & SCars & Herb19 \\
\midrule
Ground truth  
&10
&100
&100
&200
&196
&683
\\
Ours
&9
&100
&109
&231
&230
&520
\\
Error
&10\%
&0\%
&9\%
&16\%
&15\%
&28\%
\\
\bottomrule
\end{tabular}
} 

\end{table}
\begin{table*}[htb]
\footnotesize
\centering
\caption{Ablation study on the different components of our approach.}\label{tab:ablation}
\begin{tabular}{lcccccccccc}
\toprule
\multirow{2}{*}{}         & \multirow{2}{*}{ViT Backbone} & \multirow{2}{*}{Contrastive Loss} & \multirow{2}{*}{Sup. Contrastive Loss} & \multirow{2}{*}{Semi-Sup $k$-means} & \multicolumn{3}{c}{CIFAR100}                                                          & \multicolumn{3}{c}{Herbarium19}                                                       \\ 
\cmidrule(rl){6-8}
\cmidrule(rl){9-11}
                          &                      &                              &                                   &                                     & \multicolumn{1}{c}{All}  & \multicolumn{1}{c}{Old}  & New                            & \multicolumn{1}{c}{All}  & \multicolumn{1}{c}{Old}  & New                            \\ 
\midrule
\multicolumn{1}{c}{(1)} & \xmark                   & \xmark                           & \xmark                                & \xmark                                  & \multicolumn{1}{c}{34.0} & \multicolumn{1}{c}{34.8} & 32.4                           & \multicolumn{1}{c}{12.1} & \multicolumn{1}{c}{12.5} & 11.9                           \\ 
\multicolumn{1}{c}{(2)} & \cmark                  & \xmark                           & \xmark                                & \xmark                                  & \multicolumn{1}{c}{52.0} & \multicolumn{1}{c}{52.2} & 50.8                           & \multicolumn{1}{c}{12.9} & \multicolumn{1}{c}{12.9} & 12.8                           \\ 
\multicolumn{1}{c}{(3)} & \cmark                  & \cmark                          & \xmark                                & \xmark                                  & \multicolumn{1}{c}{54.6} & \multicolumn{1}{c}{54.1} & 53.7                           & \multicolumn{1}{c}{14.3} & \multicolumn{1}{c}{15.1} & 13.9                           \\ 
\multicolumn{1}{c}{(4)} & \cmark                  & \xmark                           & \cmark                               & \xmark                                  & \multicolumn{1}{c}{60.5} & \multicolumn{1}{c}{72.2} & 35.0                           & \multicolumn{1}{c}{17.8} & \multicolumn{1}{c}{22.7} & 15.4                           \\
\multicolumn{1}{c}{(5)} & \cmark                  & \cmark                          & \cmark                               & \xmark                                  & \multicolumn{1}{c}{71.1} & \multicolumn{1}{c}{\textbf{78.3}} & 56.6 & \multicolumn{1}{c}{28.7} & \multicolumn{1}{c}{32.1} & 26.9                          \\
\multicolumn{1}{c}{(6)} & \cmark                  & \cmark                          & \cmark                               & \cmark                                 & \multicolumn{1}{c}{\textbf{73.0}} & \multicolumn{1}{c}{76.2} & \textbf{66.5}                          & \multicolumn{1}{c}{\textbf{35.4}} & \multicolumn{1}{c}{\textbf{51.0}} & \textbf{27.0} \\ 
\bottomrule
\end{tabular}
\end{table*}

\begin{figure*}[t]
\begin{center}
\includegraphics[width=1.\textwidth]{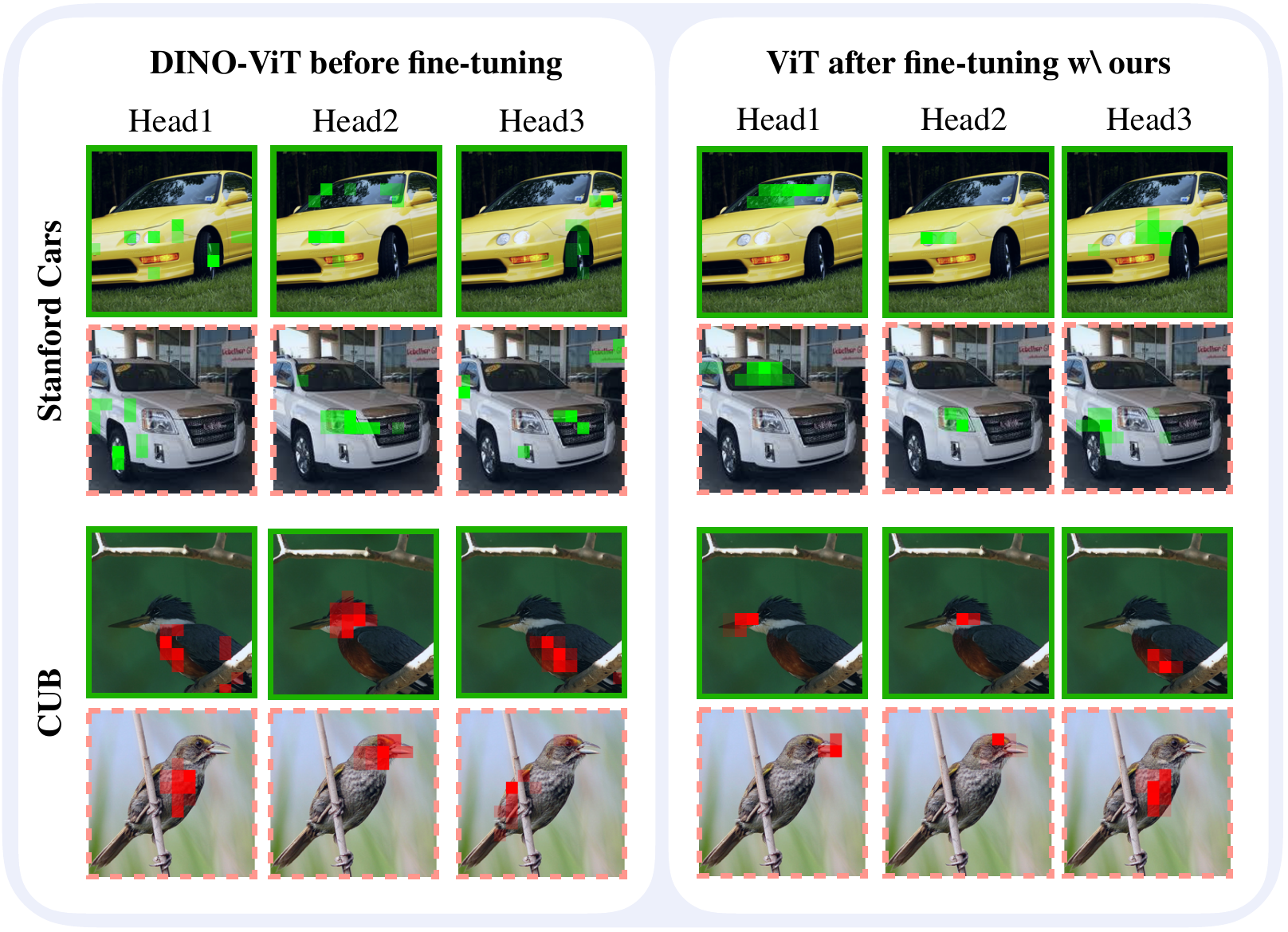}
\end{center}
\caption{Attention visualizations for the DINO-ViT model before (left) and after (right) fine-tuning with our approach. For Stanford Cars and CUB, we show an image from the `Old' (first row for each dataset) and `New' classes (second row for each dataset). Our model learns to specialize attention heads (shown as columns) to different semantically meaningful parts, which can transfer between the labelled and unlabelled categories. The model's heads learn `Windshield', `Headlight' and `Wheelhouse' for the cars, and `Beak', `Head' and 'Belly' for the birds. For both models, we select heads with as focused attention as possible. Recommended viewing in color with zoom.}
\label{fig:attention}
\end{figure*}

\subsection{Estimating the number of classes}

We report results on estimating the number of classes in~\cref{tab:k_est}.
We find that on the generic object recognition datasets, we can come very close to the ground truth number of categories in the unlabelled set, with a maximum error of 10\%.
On the fine-grained datasets, we report an average discrepancy of 18.9\%.
We note the highly challenging nature of these datasets, with many constituent classes which are visually similar.

\subsection{Ablation study}

In~\cref{tab:ablation}, we inspect the contributions of the various elements of our proposed approach.
Specifically, we identify the importance of the following components of the method: ViT backbone; contrastive fine-tuning (regular and supervised); and semi-supervised $k$-means clustering.

\paragraph{ViT Backbone} Rows (1) and (2) show the effect of the ViT model for the clustering task, as (1) and (2) represent a ResNet-50 model and ViT-B-16 trained with DINO respectively.
The ResNet model performs nearly 20\% worse aggregated over `Old' and `New' classes.
To disambiguate this from the general capacity of the architecture, note that the ImageNet linear probe discrepancy (the standard evaluation protocol for self-supervised models) is roughly 3\%~\cite{caron2021emerging}.
Meanwhile, the discrepancy in their $k$-NN accuracies on ImageNet is roughly 9\%~\cite{caron2021emerging}, suggesting why the ViT model performs so much better for the clustering task.

\paragraph{Contrastive fine-tuning} Rows (2)-(5) show the effects of introducing different combinations of contrastive fine-tuning on the target dataset.
We find that including any of the contrastive methods alone gives relatively marginal improvements over using the raw DINO features.
We find that the full benefit is only realized when combining the self-supervised and supervised contrastive losses on the target dataset.
Specifically, the combination of the contrastive losses allows us to boost aggregated clustering accuracy by a further 19\% on CIFAR100 and by 16\% on Herbarium19 (more than doubling the $ACC$ in this case).


\todo{\paragraph{Semi-supervised $k$-means} Further performance gains can be realized with semi-supervised clustering.
Across `All' classes, we observe a 2\% and 7\% increase in $ACC$ on CIFAR100 and Herbarium19 respectively.
On Herbarium19, $ACC$ on the `Old' classes is improved by 19\%.
Interestingly, it appears that semi-supervised $k$-means slightly hurts performance on the `Old' classes on CIFAR100. 
We suggest that this is an artefact of the Hungarian algorithm, which opts to assign some `clean' clusters to the `New' ground truth categories, to maximize overall $ACC$.
This can be observed in the 10\% boost provided by the semi-supervised method on the `New' classes in CIFAR100.
Furthermore, we found that if we performed the Hungarian algorithm on `Old' and `New' instances independently (allowing the reuse of clean clusters during evaluation), semi-supervised $k$-means improved $ACC$ on all data subsets.
We refer to \cref{sec:appendix_eval_metrics_details} for more details on the interaction between the reported $ACC$ and the Hungarian assignment.
}

\paragraph{Summary} Overall, we find that none of the components of our method are individually sufficient to achieve good performance across our benchmark datasets.
Specifically, the \textit{combination} of a vision transformer backbone and contrastive fine-tuning facilitates strong $k$-means clustering directly in the feature space of the model.
The semi-supervised $k$-means algorithm further allows us to guide the
clustering process with labels and achieve better $ACC$, especially on the `New' classes in the fine-grained datasets.

We further illustrate this point in the TSNE visualizations in~\cref{fig:tsne}, performed on the CIFAR10 dataset.
We show TSNE projections of raw ResNet-50 and ViT DINO features, as well as those of our model.
For the ResNet-50 features, points from the same class are generally projected close to each other, indicating that they are likely to be separable given a simple transformation (\eg a linear probe).
However, they do not form clear clusters, hinting at these features' poor down-stream clustering performance.
In contrast, the ViT features form far clearer clusters, which are further distinguished when trained with our approach.

\subsection{Qualitative results}\label{sec:qualitative_results}

Finally, we visualize the attention mechanism of our model to better understand its performance.
Specifically, in~\cref{fig:attention} we look at how the final multi-head attention layer attends to different spatial locations when supporting the output \texttt{[CLS]} token (which we use as our feature representation).
We show this both for the pre-trained DINO model and after training with our method.
We visualize the attention maps for images from the `Old' and `New' classes for Stanford Cars and CUB\@.

It is demonstrated in~\cite{caron2021emerging}  that different attention heads in the DINO model focus on different regions of an image, without the need for human annotation.
We find this to be the case, with different heads attending to disjoint regions of the image and typically focusing on important parts.
However, after training with our method, we find heads to be more specialized to semantic parts, displaying more concentrated and local attention.
In this way, we suggest the model learns to attend to a set of parts which are transferable between the `Old' and `New' classes, which allows it to better generalize knowledge from the labelled data.

\section{Conclusion}


In this paper, we have proposed a new setting for image recognition, `Generalized Category Discovery' (GCD).
We highlight three take-home messages from this work:
first, GCD is a challenging and realistic setting for image recognition;
second, GCD removes limiting assumptions in existing image recognition sub-fields such as novel category discovery and open-set recognition; 
and third, while parametric classifiers tend to overfit to labelled classes in the generalized setting, direct clustering of features from contrastively trained ViTs proves to be a surprisingly good method for classification.

\todo{\noindent\paragraph{Acknowledgements} We would like to thank Liliane Momeni for invaluable help with figures in this work.
This research is funded by a Facebook AI Research Scholarship, a Royal Society Research Professorship RP\textbackslash R1\textbackslash 191132, and the
EPSRC Programme Grant VisualAI EP/T028572/1.}

{\small\bibliographystyle{ieee_fullname}\bibliography{main}}

\clearpage

\setcounter{page}{1}

\appendix

\twocolumn[
\centering
\Large
\textbf{Generalized Category Discovery} \\
\vspace{0.5em}Appendices \\
\vspace{1.0em}
{%
\renewcommand\twocolumn[1][]{#1}%
\maketitle
\begin{center}
\centering
\captionsetup{type=figure}
\includegraphics[width=\textwidth]{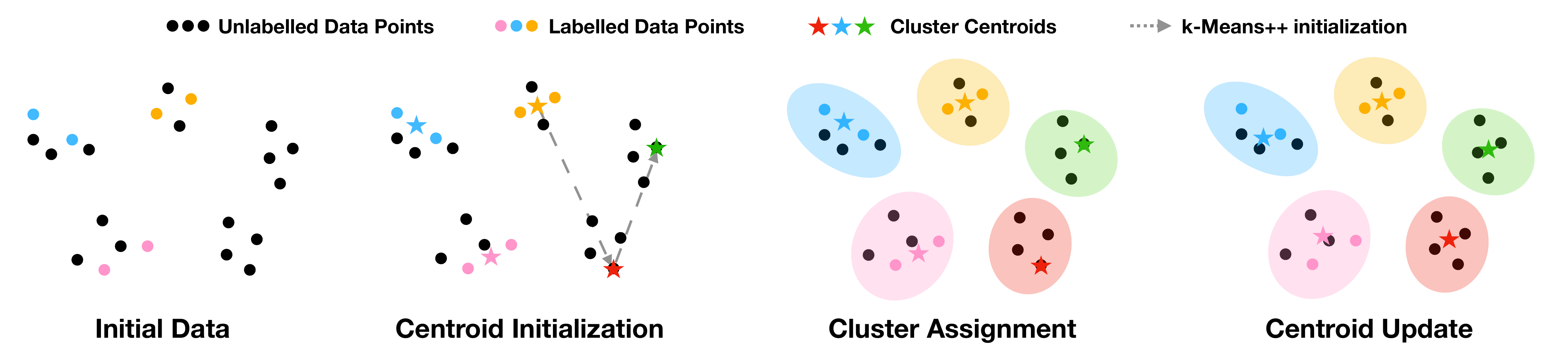}
\captionof{figure}{Semi-supervised $k$-means algorithm shown for $k = 5$. Given partially labelled data points (Initial Data), we first initialize $|\mathcal{Y_L}| = 3$ centroids by the average of labelled data points in each labelled class (shown as colored dots). Starting from these centroids, we run $k$-means++ (dashed arrows) on the unlabelled data (black dots) to further obtain $|\mathcal{Y_U} \setminus \mathcal{Y_L}| = 2$ centroids (Centroid Initialization). Having obtained $k = 5$ centroids (colored stars), we assign each data point a cluster label by identifying its nearest centroid (Cluster Assignment), after which we can update the centroids by averaging all data points in each cluster (Centroid Update). We repeat the cycle of Cluster Assignment and Centroid Update iteratively, until the $k$-means algorithm converges. During each cycle, we force the labelled data points to follow their ground-truth label, \ie all labelled points of the same class fall into the same cluster.}\label{fig:semi_sup_kmeans}
\end{center}
}]

\section{Dataset details}
\label{sec:dataset_details_sup}

Here, we describe which classes constitute the `Old' and `New' categories in the Generalized Category Discovery setting, for each dataset used in this paper. 

For all datasets, we sample 50\% of the classes as `Old' classes ($\mathcal{Y_L}$) and keep the rest as `New' ($\mathcal{Y_U} \setminus \mathcal{Y_L}$).
The exception is CIFAR100, for which we use 80 classes as `Old', following the novel category discovery literature.
For the generic object recognition datasets, we use the first $|\mathcal{Y_L}|$ classes (according to their class index) as `Old' and use the rest as `New'.
For datasets in the Semantic Shift Benchmark suite, we use the data splits provided in \cite{vaze2022openset}.
For Herbarium19, to account for the long-tailed nature of the dataset, we randomly sample the `Old' classes from the total list of classes.
Further details on the splits can be found in \cref{tab:datasets} in the main paper and in the open-source code for this project.

\section{Semi-supervised $k$-means}
\label{sec:semi_sup_kmeans_sup}

We elaborate on the semi-supervised $k$-means algorithm for GCD (from \cref{sec:semi_sup_kmeans} of the main paper) in~\cref{fig:semi_sup_kmeans}.

\section{Estimating the number of classes}

\begin{figure}[htb]
    \centering
    \includegraphics[width=\linewidth]{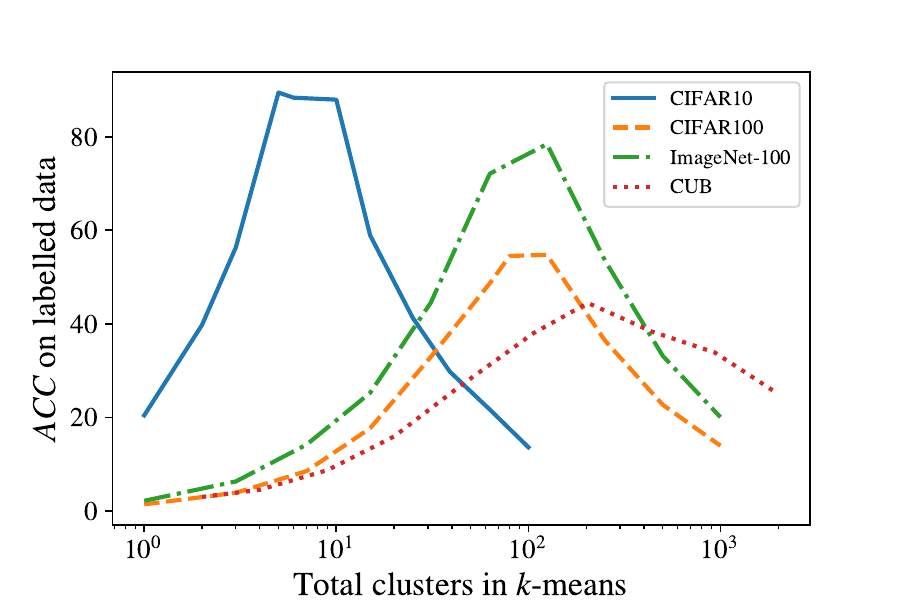}
    \caption{Plot showing how the clustering accuracy on the labelled subset ($\mathcal{D_L}$) changes as we run $k$-means, with varying $k$, on the \textit{entire} dataset ($\mathcal{D_L} \cup \mathcal{D_U}$). The ground-truth number of classes in CIFAR10, CIFAR100, ImageNet-100 and CUB are $[10, 100, 100, 200]$ respectively.}
    \label{fig:acc_k_sweep}
\end{figure}

In \cref{fig:acc_k_sweep}, we provide motivation for our algorithm in \cref{sec:k_estimation} of the main paper, which is used to estimate the number of classes in the dataset ($k = |\mathcal{Y_L} \cup \mathcal{Y_U}|$). \footnote{Note that $|\mathcal{Y_L} \cup \mathcal{Y_U}| = |\mathcal{Y_U}|$ as $|\mathcal{Y_L} \subset \mathcal{Y_U}|$. We use the former notation for clarity.}

Specifically, we plot how the clustering accuracy on the labelled subset ($\mathcal{D_L}$) changes as we run $k$-means with varying $k$ on the entire dataset ($\mathcal{D_L} \cup \mathcal{D_U})$, for a range of datasets. 
It can be seen that the $ACC$ on the labelled subset follows an approximately bell-shaped function for all datasets.
Furthermore, the function is maximized when $k$-means clustering is run with roughly the ground truth number of classes for each dataset, indicating that optimizing for this maximum is a reasonable way of identifying the total number of categories.

\section{Results on FGVC-Aircraft}
\label{sec:appendix_fgvc_aircraft}

We run the baselines and our method on the third fine-grained dataset (FGVC-Aircraft \cite{maji13fine-grained}) from the Semantic Shift Benchmark suite \cite{vaze2022openset}, reporting results in \cref{tab:appendix_fgvc_aircraft}.

\begin{table}[htb]
\footnotesize
\centering
\caption{Results on the FGVC-Aircraft \cite{maji13fine-grained} splits  from the Semantic Shift Benchmark suite \cite{vaze2022openset}.}\label{tab:appendix_fgvc_aircraft}
\begin{tabular}{lccc}
\toprule
& \multicolumn{3}{c}{FGVC-Aircraft} \\
\cmidrule(rl){2-4}
Classes       & All      & Old     & New \\\midrule
$k$-means~\cite{MackQueen67_Kmeans}
& 16.0
& 14.4
& 16.8
\\
RankStats+
& 26.9
& 36.4
& 22.2
\\
UNO+ 
& 40.3
& \textbf{56.4}
& 32.2
\\
\midrule
Ours
& \textbf{45.0}
& 41.1
& \textbf{46.9}
\\
\bottomrule
\end{tabular}
\end{table}

\section{On Hungarian assignment and clustering accuracy}
\label{sec:appendix_eval_metrics_details}
\begin{figure*}[t!]
\begin{center}
\includegraphics[width=\textwidth]{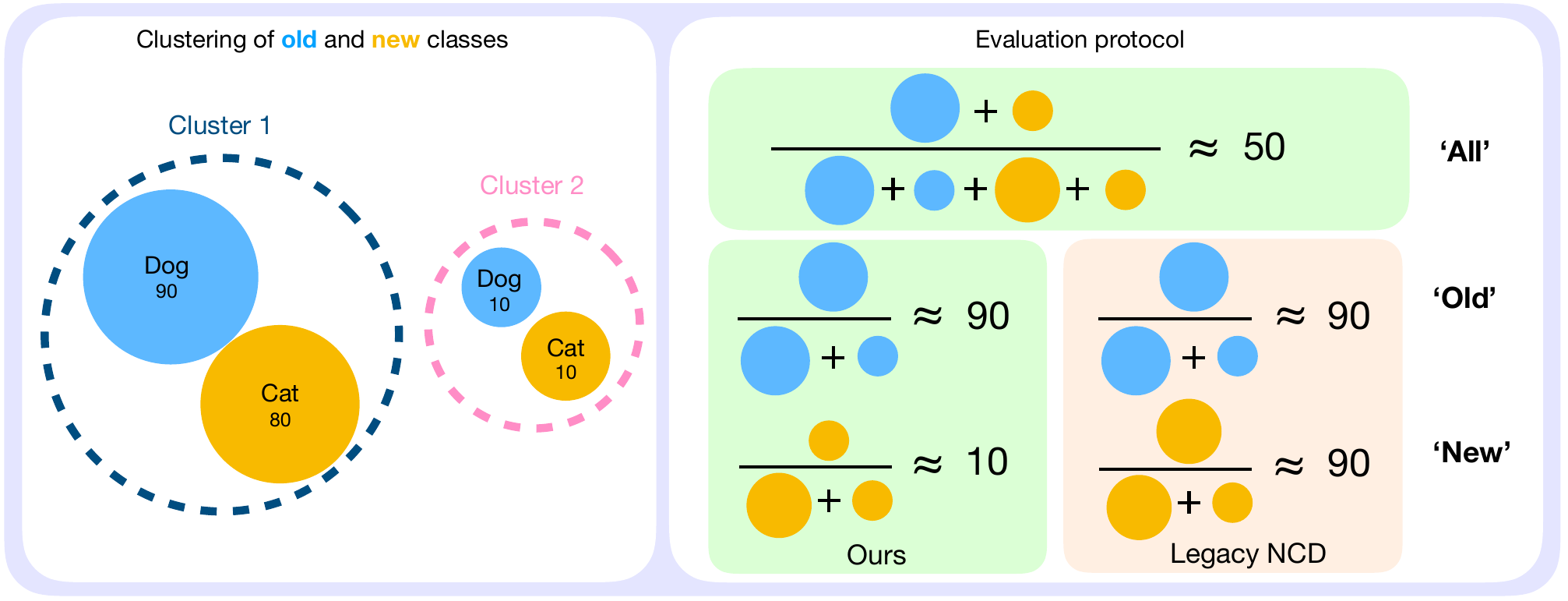}
\end{center}
\caption{An illustration of how the Hungarian algorithm affects the final clustering $ACC$. The left-hand image shows two discovered clusters (`Cluster 1' and `Cluster 2') and the ground truth labels of their constituent instances (solid colored circles). On the right, we show how $ACC$ is computed if the Hungarian assignment is computed only once (`Ours') as well as how $ACC$ is computed if the Hungarian assignment is computed independently for `Old' and `New' classes (`Legacy NCD').}
\label{fig:hungarian_and_acc}
\end{figure*}

Here, we highlight a perhaps un-intuitive interaction between how the Hungarian assignment is performed and the reported $ACC$ for all methods. 

\paragraph{Background} 
The purpose of the Hungarian algorithm \cite{kuhn1955hungarian} is to find the optimal matching between cluster indices predicted by the model and the ground truth labels.
For instance, consider a toy case with two categories $\{1, 2\}$, with an `Old' class $\mathcal{Y_L} = $\{1\} and a `New' class $\mathcal{Y_U} \setminus \mathcal{Y_L} = $\{2\}.
We further imagine the dataset to have 4 instances with ground truth labels $\{1, 1, 2, 2\}$ and a model which assigns them to clusters as $\{2, 2, 1, 1\}$. 
Note that `a cluster' here could be either a cluster as referred to in the traditional sense (e.g as in our method, a cluster of points in feature space), or simply a group of instances which are predicted the same class label by a linear classifier (as with the baselines). 
The Hungarian algorithm ensures a reported $ACC$ of 100\% in this case (as the instances have been correctly clustered together) by solving the linear assignment between ground truth labels and cluster indices as $(1\rightarrow 2, 2\rightarrow 1$). 

\paragraph{Our evaluation protocol} As stated in \cref{evaluation_protocol}, we compute the Hungarian algorithm \textit{once}, across all instances in the $\mathcal{D_U}$.
This gives us model predictions which are in the `frame of reference' of the ground truth labels for all instances.
Then, given these model predictions, we use the ground truth labels to select instances from the `Old' and `New' classes before computing the percentage of correct predictions ($ACC$) within these subsets.

\paragraph{Legacy NCD evaluation protocol} In public implementations, we find that the novel category discovery literature \cite{han21autonovel, Fini_2021_ICCV} computes $ACC$ on these subsets differently.
Specifically, the NCD literature \textit{first} uses the ground truth labels to select which instances belong to the `Old' and `New' categories, before computing the Hungarian assignment on each subset \textit{independently}.
Importantly, this allows the same discovered cluster to be used \textit{twice}. 
We suggest that this provides an overly optimistic view of model performance on the data subsets and does not quite reflect the true image recognition setting.

\paragraph{Illustration} 
The left-hand diagram in \cref{fig:hungarian_and_acc} shows two discovered clusters (`Cluster 1' and `Cluster 2') as well as the ground truth labels of their constituent instances (solid blue and yellow circles).
The blue circles indicate images from an `Old' class (e.g `Dog') and the yellow circles indicate images from a `New' one (e.g `Cat').
The radii of the circles illustrate the number of instances.

On the right, we demonstrate the `Legacy NCD' evaluation protocol for the `Old' and `New' data subsets, where the Hungarian assignment is computed twice, and independently on each data subset.
For instance, the evaluation first looks at \textit{only} `Old' instances (blue circles) and the assignment algorithm allocates the old category to Cluster 1.
The $ACC$ is then computed as the number of `Old' instances in Cluster 1 over the total number of `Old' instances.
However, subsequently, the evaluation looks \textit{only} at the `New' instances (yellow circles) and once again uses Cluster 1, assigning it to the new category.
As such, Cluster 1 is used twice, allowing the evaluation to report high $ACC$ on both data subsets.

In contrast, the Hungarian assignment can be computed over all instances, forcing the assignment of Cluster 2 to a ground truth category.
In this way, the performance on one of the data subsets is necessarily lower. 
This is how we compute $ACC$ on `Old' and `New' categories in this work, and show it as `Ours' in the figure.

Finally, we show how $ACC$ is computed over `All' categories in the unlabelled set. This protocol is followed both in this work and in the novel category discovery literature.

\section{Attention maps}

\begin{figure*}
    \centering
    \includegraphics[width=0.895\textwidth]{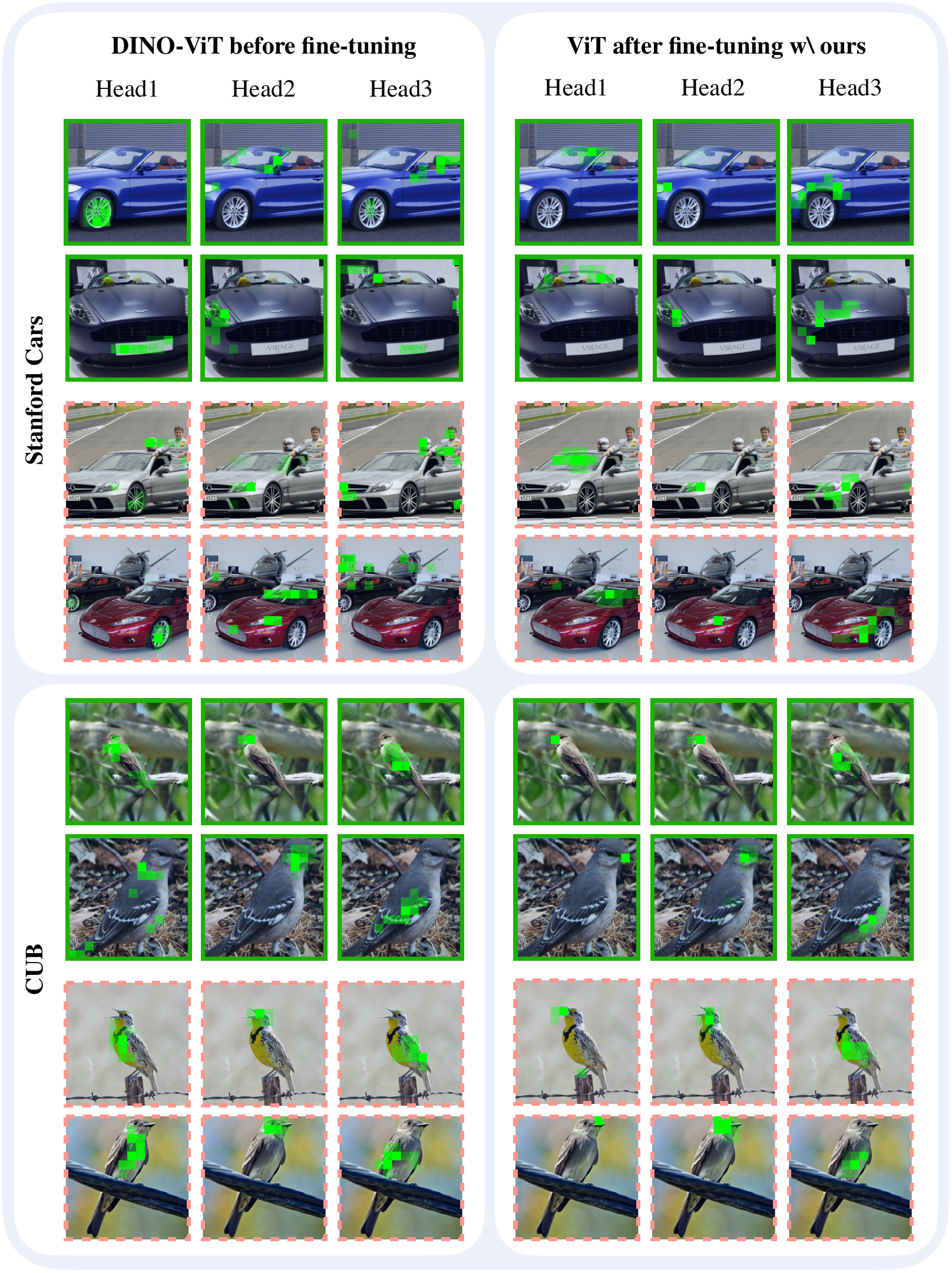}
    \caption{Attention visualizations. Attention maps for the DINO model before (left) and after (right) fine-tuning with our approach on the Stanford Cars (top) and CUB (bottom) datasets. For each dataset, we show two rows of images from the `Old' classes (solid green box) and two rows of images from the `New' classes (dashed red box). Our model learns to specialize attention heads (shown as columns) to different semantically meaningful object parts, which can transfer between the `Old' and `New' categories.}
    \label{fig:attention_viz_sup_mat}
\end{figure*}

Here, we expand upon the attention visualizations from \cref{fig:attention}.
We first describe the process for constructing them, before providing further examples in \cref{fig:attention_viz_sup_mat}.

\paragraph{Visualization construction} The attention visualizations were constructed by considering how different attention heads, supporting the output \texttt{[CLS]} token, attended to different spatial locations.
Specifically, consider the input to the final block of the ViT model, $\mathbf{X} \in \mathbbm{R}^{(HW + 1) \times D}$, corresponding to a feature for each of the $HW$ patches fed to the model, plus a feature corresponding to the \texttt{[CLS]} token.
Here, $HW = 14 \times 14 = 196$ patches at resolution $16 \times 16$ pixels.
These features are passed to a multi-head self-attention ($MHSA$) layer which can be described as:

\begin{equation}
    MHSA(\mathbf{X}) = [head_1, \dots, head_{h}]\mathbf{W_0},
\end{equation}
where 
\begin{equation}
    head_{j} = Attention(\mathbf{XW}_{j}^{Q}, \mathbf{XW}_{j}^{K}, \mathbf{XW}_{j}^{V}).
\end{equation} 
    
In other words, the layer comprises several attention heads ($h = 12$ in the ViT model) which each independently attend over the input features to the block.
We refer to \cite{vaswani2017} for more details on the self-attention mechanism.
We note that, for each feature $i \in \mathbf{X}$, an attention vector is generated for each head $j$, as
$A_{ij} \in [0, 1]^{HW + 1}$
to describe how every head $j$ relates each feature to every other feature.

We look at only the attention values for the \texttt{[CLS]} token, $A_{0j}$, and further only look at the elements which attend to spatial locations.
We find that, while some heads have uninterpretable attention maps, certain heads specialize to attend to coherent semantic object parts.

\paragraph{Discussion on attention visualizations}

We provide further attention visualizations in \cref{fig:attention_viz_sup_mat}. We show the \textit{same} attention heads as shown in \cref{fig:attention} (both for models trained with our method and for the original DINO model). 

We first note that the DINO features often attend to salient object regions.
For instance, `Head 1' of the model often focuses on the wheel of the car (in the Stanford Cars examples), while `Head 2' of the model generally attends to the heads of the birds (in the CUB examples).
Overall, however, with the pre-trained DINO model, there is relatively little semantic consistency between the attention maps of a given head (\ie within columns for each dataset). 

In contrast, we find that the models trained with our approach specialize attention heads in semantically meaningful ways.
The heads shown correspond to `Windshield', `Headlight' and `Wheelhouse' for the Stanford Cars model, and `Beak', `Head' and `Belly' for the CUB model.
We find these maps to be relatively robust to nuisance factors such as pose and scale shift, as well as distracting objects in the image. 
We note a failure case (rightmost image, Row 2), as the `Wheelhouse' attention head is forced to attend to miscellaneous regions of the car, as the car's wheelhouse is not visible in this image.

The ability of the model to identify and distinguish different semantic parts of an object is useful for the GCD task. 
Particularly in the fine-grained setting, the constituent set of parts of an object (`Head', `Beak', `Belly' \etc for the birds) transfer between `Old' and `New' classes.
Thus, we suggest that the attention mechanism of the model allows it to generalize its understanding from the labelled `Old' classes, and apply it to the unlabelled `New' ones.

\section{Broader impact and limitations}

Our method assigns labels to images in an unsupervised manner, including discovering new labels.
Even more than standard image classification methods, it should be used with care (e.g., by manually checking the results) in sensitive contexts, such as processing personal data.

We also note some practical limitations.
First, we assume that there is no domain shift between the labelled and unlabelled subsets.
For instance, we are not tackling the problem of a single model reliably classifying photographs and paintings of the same classes. 
Second, we do not consider the streaming setting (also known as continual learning):
one would need to re-train the model from scratch as new data becomes available.

As for the data used in the experiments, we use standard third-party datasets in a manner compatible with their licenses.
Some of these datasets contain Internet images that feature, often incidentally, people  --- see~\cite{prabhu20large,yang21a-study,asano21pass} for an in-depth discussion of the privacy implications.

\end{document}